\documentclass{article}

\usepackage{amsmath}
\usepackage{amssymb}
\usepackage{amsfonts}
\usepackage{bbm}

\usepackage{float}
\usepackage{algorithm}
\usepackage{graphicx}
\usepackage{subcaption}

\renewcommand{\mathbf}{\boldsymbol} 
\newcommand{\mb}{\mathbf}
\newcommand{\mc}{\mathcal}

\newcommand{\bx}{\mb{x}}
\newcommand{\bw}{\mb{w}}

\newcommand{\bE}{\mb{E}}
\newcommand{\bZ}{\mb{Z}}

\newcommand{\bmu}{\mb{\mu}}
\newcommand{\bSigma}{\mb{\Sigma}}
\def\b0{\mb{0}}

\newcommand{\DS}{\Delta_{\bSigma}}
\newcommand{\Ds}{\Delta_{\sigma}}

\newcommand{\cX}{\mc{X}}
\newcommand{\cY}{\mc{Y}}
\newcommand{\cB}{\mc{B}}
\newcommand{\cW}{\mc{W}}

\newcommand{\cN}{\mc{N}}
\newcommand{\cS}{\mc{S}}

\newcommand{\cD}{\mc{D}}

\newcommand{\RN}{R_{\cN}}

\newcommand{\RS}{R_{\cS}}

\newcommand{\EX}{\mathbb{E}}

\newcommand{\bV}{\mathbb{V}}

\newcommand{\eps}{\epsilon}

\def\el2{\ell_2}

\def\bell{\mb{\ell}}

\newcommand{\paren}[1]{ \left( #1 \right) }
\newcommand{\brac}[1]{\left[ #1 \right]}
\newcommand{\Brac}[1]{\left\lbrace #1 \right\rbrace}

\newcommand{\set}[1]{\left\{ #1 \right\}}
\newcommand{\norm}[2]{\left\| #1 \right\|_{#2}}
\newcommand{\magnitude}[1]{\left| #1 \right|}

\def\nn{\nonumber}

\title{Risk Bounds for Low Cost Bipartite Ranking}
\author{San Gultekin and John Paisley}
\date{\today}
\begin{document}

\maketitle	
	
\begin{abstract}
	Bipartite ranking is an important supervised learning problem; however, unlike regression or classification, it has a quadratic dependence on the number of samples. To circumvent the prohibitive sample cost, many recent work focus on stochastic gradient-based methods. In this paper we consider an alternative approach, which leverages the structure of the widely-adopted pairwise squared loss, to obtain a stochastic and low cost algorithm that does not require stochastic gradients or learning rates. Using a novel uniform risk bound---based on matrix and vector concentration inequalities---we show that the sample size required for competitive performance against the all-pairs batch algorithm does not have a quadratic dependence. Generalization bounds for both the batch and low cost stochastic algorithms are presented. Experimental results show significant speed gain against the batch algorithm, as well as competitive performance against state-of-the-art bipartite ranking algorithms on real datasets.  
\end{abstract}

\section{Introduction}
\label{sec:intro}

Binary classification is among the most widely studied machine learning problems, with many applications. Given a binary labeled dataset, the aim is to learn a mapping from the features to the labels. The performance of a learning algorithm is typically gauged in terms of classification error. However, in situations such as cost-sensitive learning \cite{McMahan_2013} and imbalanced learning \cite{Hu_2016, Hong_2007} this choice may not be appropriate. For example, in online advertising \cite{McMahan_2013} one is typically concerned with separating the interesting ads from the rest. This problem is also known as bipartite ranking, where the aim is to rank the ``positive'' inputs higher than the ``negative'' ones.

\subsection{Problem Setup}
Let $\cX$ be a $D$-dimensional input domain and $\cY$ the label domain. The bipartite ranking problem is defined for binary labeled samples, $\cY = \{0,1\}$. These sample pairs are generated i.i.d. from an unknown distribution $\cD$ on $\cX \times \cY$. A ranking function $f:\cX \rightarrow \mathbb{R}$ is a mapping from the inputs to a scalar-valued score. In this paper we are interested in linear ranking functions of the form $f(\bx)=\bw^\top \bx$ with parameter vector $\bw$. The goal in bipartite ranking is to assign higher scores to the inputs with label 1. Let $\bx^1$ and $\bx^0$ denote two samples with corresponding labels of $1$ and $0$. In many ranking settings we are given a set of features that encodes the preference of one sample over another \cite{McMahan_2013}, which can be represented by $\Psi(\bx^1, \bx^0)$. It is also possible that the individual features $\bx_0$ and $\bx_1$ are not defined explicitly; for example when a single user is shown two ads and prefers one over another. Given this, a widely used performance metric is the Wilcoxon-Mann-Whitney statistic
\begin{align}\label{eq:l_wmw}
\bell_{\text{WMW}}(\bw,\Psi(\bx^1,\bx^0)) = \mathbb{I}\Brac{\bw^\top \Psi(\bx^1, \bx^0) > 0} + \frac{1}{2} \mathbb{I}\Brac{\bw^\top  \Psi(\bx^1, \bx^0) = 0} ~.
\end{align}
Based on this we define the risks
\begin{align}\label{eq:auc_risks}
R^\text{\tiny AUC}(\bw) = \EX_{\substack{x^1 \sim D^1 \\ x^0 \sim D^0}}\brac{\bell_{\text{WMW}}(\bw,\Psi(\bx^1,\bx^0))} ,~~  R^\text{\tiny AUC}_\cN(\bw) = \frac{1}{N_1 N_0} \sum_{i=1}^{N_1} \sum_{j=1}^{N_0} \bell_{\text{WMW}}(\bw,\Psi(\bx_i^1,\bx_j^0)),
\end{align}
where $R^{\text{\tiny AUC}}(\bw)$ corresponds to the actual AUC risk
and is obtained by taking expectation over the class-conditional distributions $D^i = D(\bx|y=i)$. Let the sample set $\cN=\{\bx_1^0,\ldots,\bx_{N_0}^0,\bx_1^1,\ldots,\bx_{N_1}^1\}$, containing $N_0$ and $N_1$ samples with labels 0 and 1, respectively, and define $N:=N_0+N_1$. This gives the empirical risk $R^\text{\tiny AUC}_\cN(\bw)$ to be minimized in practice. The main drawback of this approach is that the objective function is now a sum of indicator functions, an NP-hard problem. A widely-used approach to handle this problem is to replace the intrinsic loss function $\bell_{\text{WMW}}$ with a convex one, often written as $\bell_\phi$ \cite{Zhao_2011,Gao_2013}. 

In this paper we are interested in the pairwise squared loss as it is a consistent estimator of AUC \cite{Gao_2015}, and widely preferred by recent work \cite{Boissier_2016, Ding_2015, Kar_2013}:
\begin{align}
\bell_{\phi}(\bw,(\bx^1,\bx^0)) = \frac{1}{2} \brac{1-\bw^\top\Psi(\bx_i^1,\bx_j^0)}^2,
\end{align}
which produces its own corresponding actual and empirical $\phi$-risks that parallels Eq.\ \eqref{eq:auc_risks}. 
Let $\bSigma_N = 1/(N_1 N_0)\sum_{i=1}^{N_1} \sum_{j=1}^{N_0} \Psi(\bx_i^1,\bx_j^0) \Psi(\bx_i^1,\bx_j^0)^\top$ and $\bmu_N = 1/(N_1 N_0)\sum_{i=1}^{N_1} \sum_{j=1}^{N_0} \Psi(\bx_i^1,\bx_j^0)$. The empirical risk for this problem can then also written as the optimization of $R_{\cN}^\phi(\bw) = (1/2) \bw^\top \bSigma_N \bw - \bmu_N^\top \bw$. For what follows we will drop the $\phi$ symbol and refer to $\bell_\phi$, $R^\phi$, and $R_{\cN}^\phi$ simply as $\bell$, $R$, and $\RN$.

The focus of this paper is on two algorithms: The first one is a Batch Bipartite Ranking (BBR) algorithm (Algorithm 1), which aims at minimizing $R_\cN$ based on all pairs available in $\cN$. We provide a new theoretical analysis of BBR in Section \ref{sec:bbr}. Due to the quadratic growth of sample size, $O(N_1  N_0)$, the sample cost of BBR quickly becomes prohibitive. Therefore we also propose a new Low Cost Bipartite Ranking (LCBR) algorithm (Algorithm 2), which, given the same sample set $\cN$, {subsamples} $S$ pairs uniformly at random with replacement. The main goal of this paper is to analyze the subsample size $S$ required for LCBR to be competitive with BBR. As we show in Section \ref{sec:lcbr}, $S$ does not have such quadratic dependence. Section \ref{sec:rel} discusses related work. We show experiments in Section \ref{sec:exp} and conclude in Section \ref{sec:con}.

\begin{minipage}[t!]{.5\linewidth}
	\begin{algorithm}[H] \label{alg:bbr}
		\caption{BBR}
		\textbf{Input:~~~} Sample set $\cN$ \\
		\phantom{\textbf{Input:~~~}} Regularization parameter ($W_*$) \\
		\textbf{Output:} Linear ranker's weight $\bw_N$ \\
		1. Initialize $\bmu_N \gets \b0$, $\bSigma_N \gets \b0$ \\
		2. \texttt{//Accummulation} \\
		3. \textbf{for} $i = 1,\ldots,N_1$ \\
		4. $~$ \textbf{for} $j = 1,\ldots,N_0$ \\
		5. $~~$ $\bmu_N \gets \bmu_N + \frac{1}{N_1 N_0}(\bx_i^1 - \bx_j^0)$ \\
		6. $~~$ $\bSigma_N \gets \bSigma_N + \frac{1}{N_1 N_0}(\bx_i^1 - \bx_j^0)(\bx_i^1 - \bx_j^0)^\top$ \\
		7. $~$ \textbf{end for} \\
		8. \textbf{end for} \\
		9. \texttt{//Empirical Risk Minimization} \\
		10. $\bw_N \gets \underset{\bw \in B_2(W_*)}{\arg\min} ~~ \frac{1}{2}\bw^\top \bSigma_N \bw - \bmu_N^\top \bw$
	\end{algorithm}
\end{minipage}%
\begin{minipage}[t!]{.5\linewidth} 
	\begin{algorithm}[H]
		\caption{LCBR}
		\textbf{Input:~~~} Sample set $\cN$, subsample size ($S$) \\
		\phantom{\textbf{Input:~~~}} Regularization parameter ($W_*$) \\
		\textbf{Output:} Linear ranker's weight $\bw_S$ \\
		1. Initialize $\bmu_S \gets \b0$, $\bSigma_S \gets \b0$ \\
		2. \texttt{//Accummulation} \\
		3. \textbf{for} $s = 1,\ldots,S$ \\
		4. $~$ Sample $(i_s,j_s)$ uniformly with replacement \\
		5. $~$ $\bmu_S \gets \bmu_S + \frac{1}{S}(\bx_{i_s}^1 - \bx_{j_s}^0)$ \\
		6. $~$ $\bSigma_S \gets \bSigma_S + \frac{1}{S}(\bx_{i_s}^1 - \bx_{j_s}^0)(\bx_{i_s}^1 - \bx_{j_s}^0)^\top$ \\
		7. \textbf{end for} \\
		8. \phantom{\textbf{end for}} \\
		9. \texttt{//Empirical Risk Minimization} \\
		10. $\bw_S \gets \underset{\bw \in B_2(W_*)}{\arg\min} ~~ \frac{1}{2}\bw^\top \bSigma_S \bw - \bmu_S^\top \bw$
	\end{algorithm}
\end{minipage}%

\textbf{Remark:} At this point, for the ease of presentation we focus on the case $\Psi(\bx_i^1,\bx_j^0) = \bx_i^1 - \bx_j^0$ (which is reflected in Algorithms 1 and 2). All the theory developed in this paper is valid for the most general case of $\Psi(\bx_i^1,\bx_j^0)$, however. In particular, imposing a norm bound on the individual features $\bx_i^1, \bx_j^0$ is equivalent to imposing a scaled norm bound on $\Psi(\bx_i^1,\bx_j^0)$. 

\textbf{Additional assumptions:} We assume the input domain is compact, $\cX \subseteq \cB_2(X_*)$ which implies $\norm{\bx}{2} \leq X_*$, and that the domain of ranking functions is compact, taking $\cW = \cB_2(W_*)$ such that $\norm{\bw}{2} \leq W_*$. Here $\cB_{p}(r) := \{\bx \in \cX: \norm{\bx}{p} \leq r\}$ is the $\ell_p$-ball of radius $r$. The first assumption is related to data preprocessing where the features are scaled appropriately; the second assumption corresponds to regularization. For both cases we chose the $\ell_2$-norm as it provides a dimension-independent upper bound; however it is still possible to derive bounds using other norms. Let $\bw_N$ and $\bw_S$ be the minimizer of the empirical risk objectives (Algorithms 1 and 2, Line 10). Also for clarity we focus on the case where $\bSigma, \bSigma_N, \bSigma_S \succ \b0$ in the sequel.\footnote{Where we defined $\bSigma = \EX_{\bx_1 \sim P_1, \bx_0 \sim P_0}[(\bx_1-\bx_0)(\bx_1-\bx_0)^\top]$.}  This means $\bw_\star$, $\bw_N$, and $\bw_S$ have unique values. Note that, this last assumption is only for presentation  purposes and the results derived in this paper apply to the most general case where $\bSigma, \bSigma_N, \bSigma_S \succeq \b0$.

\section{Batch Bipartite Ranking (BBR)}
\label{sec:bbr}

In order to analyze the more efficient LCBR, we first derive a risk bound for the corresponding BBR. The risk bounds derived in this paper are with respect to the best-in-class ranking function. For the sample set $\cN$ we define $\rho := N_1/(N_0+N_1)$ as the label skew. Also, for the unit sphere $\cS^{D-1}$ let $C_\cS(\eps)$ denote the covering number based on $\ell_2$-balls of radius $\eps.$ The main result of this section is the following risk bound based on the metric entropy of $\cW$.

\textbf{Theorem 1.} For a given sample set $\cN$, define $\bw_N:=\arg\min_{\bw \in \cW} \RN(\bw)$ and $\bw_\star = \arg\min_{\bw \in \cW} R(\bw)$. Also define the constants $C_1:=8 X_*^2 W_* + 4X_*$ and $C_2:=3 X_*^2 W_*^2 + 2 X_* W_*$. Then
\begin{align}\label{eq:bbr_regret}
\underset{\cN \sim \cD^N}{P} \bigg( R(\bw_N)-R(\bw_\star) \geq \eps \bigg) \leq 2 ~ C_{\cS}\paren{\frac{\eps}{4C_1}} ~ \exp \Brac{-\frac{\eps^2}{8 C_2^2}\rho (1-\rho) N} ~.
\end{align}
Here we used the shorthand $\cD^N$ to denote the product measure over the samples, $[\cD^1]^{N_1} \otimes [\cD^0]^{N_0}$ where $[\cD^i]^{N_i} = \otimes_{i=1}^{N_i} \cD^i$. This result is comparable to the bound given for linear regression based on covering numbers \cite{Mohri_2012}. One distinction, however, is the dependence on skew $\rho(1-\rho)$; when $\rho$ is close to 0 or 1, the learner requires significantly more samples to achieve the same bound. The exponential term in Eq.\ \eqref{eq:bbr_regret} is comparable to the one obtained for AUC loss in Theorem 5 of \cite{Agarwal_2005}. On the other hand, when $\rho(1-\rho)=O(1)$ and $N$ is large enough to bound the covering number by the exponential term in Eq.\ \eqref{eq:bbr_regret} we get the typical rate $O(\sqrt{\log(1/\delta)/N})$.

As the rate in Eq.\ \eqref{eq:bbr_regret} depends on $N_1+N_0$ and not $N_1 N_0$, it is natural to only consider pairs of independent samples instead of their Cartesian product, which reduces the analysis to that of linear regression. However, this is not done in practice as it would discard information \cite{Boissier_2016}. Indeed, for this reason, a number of works consider the all-pair problem, e.g. \cite{Boissier_2016, Zhao_2011, Gao_2013}; we do so similarly. Below, we prove Theorem 1 using the following Lemma.

\textbf{Lemma 1.~} For a given sample set $\cN$ and constants $C_1:=8 X_*^2 W_* + 4X_*$, $C_2:=3 X_*^2 W_*^2 + 2 X_* W_*$,
\begin{align}\label{eq:bbr_uc}
\underset{\cN \sim \cD^N}{P} \bigg( \underset{\bw \in \cW}{\sup} \magnitude{ R(\bw)-\RN(\bw) } \geq \eps \bigg) \leq 2 ~ C_{\cS}\paren{\frac{\eps}{2C_1}} ~ \exp \Brac{-\frac{\eps^2}{2 C_2^2}\rho (1-\rho) N} ~.
\end{align}

\textbf{Proof.~} Recall that pairwise squared loss is defined as $\bell(\bw,\bx^1,\bx^0) = (1/2) (1-h_{\bw}(\bx^1,\bx^0))^2$ for $h_{\bw}(\bx^1,\bx^0) = \bw^\top (\bx^1-\bx^0)$. Also define $\Phi_\cN(\bw) := R(\bw)-\RN(\bw)$. For $\bw_1,\bw_2 \in \cW$,
\begin{align}
\bigg| \ell(\bw_1,\bx^1,\bx^0)&-\ell(\bw_2,\bx^1,\bx^0) \bigg| = \left| \frac{1}{2}(1-h_{\bw_1}(\bx^1,\bx^0))^2 - \frac{1}{2}(1-h_{\bw_2}(\bx^1,\bx^0))^2 \right| \nn \\
&\quad\quad \leq \frac{1}{2} \big| 2 - h_{\bw_1}(\bx^1,\bx^0) - h_{\bw_2}(\bx^1,\bx^0) \big| ~ \big| h_{\bw_2}(\bx^1,\bx^0) - h_{\bw_1}(\bx^1,\bx^0) \big| \nn \\
&\quad\quad \leq (2X_* W_* + 1) \magnitude{(\bw_2-\bw_1)^\top(\bx^1-\bx^0)} \nn \\
&\quad\quad \leq (4X_*^2 W_* + 2 X_*) \norm{\bw_1-\bw_2}{2} ~.
\end{align}
Using this bound we have
\begin{align}
&\magnitude{ \Phi_\cN(\bw_1) - \Phi_\cN(\bw_2) } = \left| R(\bw_1) - \RN(\bw_1) - R(\bw_2) + \RN(\bw_2) \right| \nn \\
&~~ \leq \bigg| \EX_{\substack{x^1 \sim D^1 \\ x^0 \sim D^0}}\left[ \ell(\bw_1,\bx^1,\bx^0)-\ell(\bw_2,\bx^1,\bx^0)\right] \bigg| + \frac{1}{N_1 N_0} \sum_{i=1}^{N_1} \sum_{j=1}^{N_0} \bigg| \ell(\bw_1,\bx_i^1,\bx_j^0)-\ell(\bw_2,\bx_i^1,\bx_j^0)\bigg| \nn \\
&~~ \leq (8X_*^2 W_* + 4 X_*) \norm{\bw_1-\bw_2}{2}.
\end{align}

We have shown that $\magnitude{\Phi_\cN(\bw_1)-\Phi_\cN(\bw_2)} \leq C_1 \norm{\bw_1-\bw_2}{2}$. Let $\{\cB_i\}_{i=1}^{I}$ be a set of $\ell_2$-balls of radius $\eps$ covering $\cW$. Then
\begin{align}
P_\cN \paren{\sup_{\bw \in \cW} \magnitude{\Phi_\cN(\bw)} \geq \eps} \leq \sum_{i=1}^I P_\cN \paren{ \sup_{\bw \in \cB_i} \magnitude{\Phi_\cN(\bw)} \geq \eps } \leq \sum_{i=1}^I P_\cN \paren{ \magnitude{\Phi_\cN(\bw_i)} \geq \eps/2 } . 
\end{align}
Now that the weight vector and samples are decoupled, we can bound the deviation of $\Phi_\cN(\bw)$ for a fixed $\bw$. First note that $\EX[\Phi_\cN(\bw)] = 0$ by definition of $R(\bw)$ and $\RN(\bw)$. Now consider perturbation of a single variable---define $\cN' = (\bx^1,\ldots,\bx',\ldots,\bx^{N})$ which matches $\cN$ everywhere except $\bx'$. We have two cases: 
\begin{enumerate}
	\item[(i)] When $\bx'$ has corresponding label $0$, we have
	\begin{align}
	|\Phi_\cN(\bw)-\Phi_{\cN'}(\bw)| &=\footnotesize \magnitude{\frac{1}{2N_0 N_1} \sum_{i=1}^{N_1} [1-\bw^\top(\bx_i^1 - \bx_j^0)]^2 - \frac{1}{2N_0 N_1} \sum_{i=1}^{N_1} [1-\bw^\top(\bx_i^1 - \bx')]^2} \nn \\
	&\leq \frac{1}{2N_0 N_1} N_1 6 X_*^2 W_*^2 + \frac{1}{2N_0 N_1} N_1 4 X_* W_*
	\end{align}
	From this last line it follows that $\magnitude{\Phi_\cN(\bw)-\Phi_{\cN'}(\bw)} \leq C_2/N_0$.
	
	\item[(ii)]  Similarly when $\bx'$ has corresponding label $1$: $\magnitude{\Phi_\cN(\bw)-\Phi_{\cN'}(\bw)} \leq C_2/N_1$.
\end{enumerate}
As the differences are bounded and the inputs are independent, applying McDiarmid's inequality yields $P(\magnitude{\Phi_\cN(\bw)}\geq\eps/2) \leq 2 \exp\{-(\eps^2/2C_2^2)\rho(1-\rho)N\}$. This, along with $I=C_\cS(\eps/(2C_1))$, implies the bound in Lemma 1. $\hfill \square$

\textbf{Proof of Theorem 1.~} By definition of $\bw_\cN$ and $\bw_\star$ we have $R(\bw_N)-R(\bw_\star) \geq 0$. Also,
\begin{align}
R(\bw_N)-R(\bw_{\star}) &= [R(\bw_N) - \RN(\bw_N) + \RN(\bw_{\star}) - R(\bw_{\star})] + [\RN(\bw_N) - \RN(\bw_{\star})] \nn \\
&\leq \magnitude{R(\bw_N)-\RN(\bw_N)} + \magnitude{R(\bw_{\star})-\RN(\bw_{\star})} ~.
\end{align}
where the second line follows from $\RN(\bw_N) - \RN(\bw_{\star}) \leq 0$ and the triangle inequality. We next bound both terms by $\eps/2$, and according to Lemma 1, the bounds hold simultaneously with probability at least $1-C_{\cS}(\eps/4C_1) 2\exp \{-(\eps^2/(8 C_2^2))\rho (1-\rho) N\}$. $\hfill \square$

\section{Low Cost Bipartite Ranking (LCBR)}
\label{sec:lcbr}

In this section we derive risk bounds for LCBR, a subsampling strategy for approximately optimizing BBR. Our main goal is to obtain a bound similar to that in Eq.\ \eqref{eq:bbr_regret}. We start with a brief comparison of LBCR and BBR: In Section \ref{sec:intro} we noted that the empirical risk objective of BBR can be written as $R_{\cN}(\bw) = (1/2) \bw^\top \bSigma_N \bw - \bmu_N^\top \bw$ where $\bSigma_N = 1/(N_1 N_0) \sum_{i=1}^{N_1} \sum_{j=1}^{N_0} (\bx_i^1-\bx_j^0) (\bx_i^1-\bx_j^0)^\top$ and $\bmu_N = 1/(N_1 N_0) \sum_{i=1}^{N_1} \sum_{j=1}^{N_0} (\bx_i^1-\bx_j^0)$. Here $\bSigma_N$ and $\bmu_N$ are constructed using all $N_1 N_0$ pairs available. On the other hand, LBCR {subsamples} $S$ \textit{pairs} from the {fixed} $N_1 N_0$ \textit{total pairs} uniformly at random with replacement. The pairs obtained this way are denoted by the set $\cS = \{(x_1^1,x_1^0),\ldots,(x_S^1,x_S^0)\}$. It can be seen that the elements of $\cS$ are random variables sampled from a uniform distribution {conditional} on $\cN$. Thus, while the elements of $\cN$ are sampled from the class-conditionals, i.e. $\cD^i$, the elements of $\cS$ are sampled from the uniform distribution $\cD(\cN)$. With a slight abuse of notation, we will denote this by $(\bx_i^1,\bx_j^0) \sim \cN$. The corresponding objective of LCBR is $R_{\cS}(\bw) = (1/2) \bw^\top \bSigma_S \bw - \bmu_S^\top \bw$ with $\bmu_S$ and $\bSigma_S$ the first and second moments computed on the subsample. 


In order to derive a risk bound for $|R(\bw_S)-R(\bw_\star)|$ we first need to bound $|\RN(\bw_N)-\RN(\bw_S)|$. However, in the latter expression, the weight vectors and the samples are not independent. For this reason, we again use a uniform convergence argument. Here we use matrix and vector concentration to obtain the bounds necessary. In particular, the following lemma provides concentration inequalities for two key variables.

\textbf{Lemma 2.~} For $\bw_1, \bw_2 \in \cW$, define $\DS := \bw_1^\top (\bSigma_S - \bSigma_N) \bw_1$ and $\Ds := (\bw_1 - \bw_2)^\top (\bmu_N-\bmu_S)$. The following hold:

\qquad ~(i) $P(\sup_{\bw_1 \in \cW} \magnitude{\DS} \geq \eps) \leq 2D \exp \{-S \eps^2 / (8 \norm{\bSigma_N}{2} X_*^2 W_*^4 + (16/3) \eps X_*^2 W_*^2)\}$

\qquad (ii) $P(\sup_{\bw_1,\bw_2 \in \cW} \magnitude{\Ds} \geq \eps) \leq 2 \exp \{-1/2 (\eps\sqrt{S}/(2X_*W_*)-1)^2\}$

The proof is given in the appendix. We can now bound the difference of the empirical risks with high probability.

\textbf{Theorem 2.~} For a given sample set $\cN$ and its subsample set $\cS$ let $\bw_N = \arg\min_{\bw \in \cW} \RN(\bw)$ and $\bw_S = \arg\min_{\bw \in \cW} \RS(\bw)$. If the subsample size satisfies

\begin{align}\label{eq:subsample_comp}
S \geq \max \biggl\{ \log(4D/\delta) \frac{\norm{\bSigma_N}{2}X_*^2 W_*^4 + (1/3)\eps X_*^2 W_*^2}{\eps^2/32} ~,~ \frac{X_*^2 W_*^2}{\eps^2/4} \brac{\sqrt{2\log(4/\delta)}+1}^2 \biggr\}
\end{align}
then
\begin{align}
\underset{\cS \sim \cN}{P} \bigg( |\RN(\bw_S)-\RN(\bw_N)| \geq \eps \bigg) \leq 1-\delta ~.
\end{align}

\textbf{Proof.~} We define the following pointwise difference
\begin{align}
\Delta(\bw):= \RS(\bw) - \RN(\bw) = \frac{1}{2} \bw^\top (\bSigma_S - \bSigma_N) \bw + \bw^\top (\bmu_N-\bmu_S) ~.
\end{align} 
Our first task is to show that the following inequality holds,
\begin{align}
0 \leq \RN(\bw_S) - \RN(\bw_N) \leq \Delta(\bw_N) - \Delta(\bw_S) ~.
\end{align}
The LHS of this inequality holds by the definition of $\bw_N$ and $\bw_S$. For the RHS we have
\begin{align}
\RN(\bw_S) - \RN(\bw_N) = \Delta(\bw_N) - \Delta(\bw_S) - [\RS(\bw_N)-\RS(\bw_S)] \leq \Delta(\bw_N) - \Delta(\bw_S)
\end{align}
since $\RS(\bw_N)-\RS(\bw_S) \geq 0$. It is therefore sufficient prove a high probability bound for $\magnitude{\Delta(\bw_N) - \Delta(\bw_S)}$. Next,
\begin{align}
\magnitude{\Delta(\bw_N) - \Delta(\bw_S)} & =\footnotesize \magnitude{\frac{1}{2}\bw_N^\top (\bSigma_S-\bSigma_N) \bw_N - \frac{1}{2}\bw_S^\top (\bSigma_S-\bSigma_N) \bw_S + (\bw_S-\bw_N)^\top (\bmu_N-\bmu_S)} \nn \\
& \leq \footnotesize \biggl| \frac{1}{2}\bw_N^\top (\bSigma_S-\bSigma_N) \bw_N - \frac{1}{2}\bw_S^\top (\bSigma_S-\bSigma_N) \bw_S \biggr| + \biggl| (\bw_S-\bw_N)^\top (\bmu_N-\bmu_S) \biggr|.
\end{align}
We bound each of these terms by $\epsilon/2$ with probability at least $1-\delta/2$.
Note that both terms have weight vectors coupled with samples so we cannot apply concentration inequalities directly. However, they can be upper bounded by the expressions in Lemma 2. In particular, for the quadratic term $\magnitude{\frac{1}{2}\bw_N^\top (\bSigma_S-\bSigma_N) \bw_N - \frac{1}{2}\bw_S^\top (\bSigma_S-\bSigma_N) \bw_S} \leq \sup_{\bw_1,\bw_2 \in \cW} \magnitude{\DS}$. We then apply Lemma 5(i) with threshold $\eps/2$ and probability $\delta/2$ which yields the first term in Eq.\ \eqref{eq:subsample_comp}. For the linear term, $\magnitude{(\bw_S-\bw_N)^\top (\bmu_N-\bmu_S)} \leq \sup_{\bw_1,\bw_2 \in \cW} \magnitude{\Ds}$. Applying Lemma 5(ii) with threshold $\eps/2$ and probability $\delta/2$ yields the second term. $\hfill \square$

Theorem 2 shows that the subsample size $S$ required to decrease the difference $\magnitude{\RN(\bw_S)-\RN(\bw_N)}$ does not depend on the number of total pairs $N_1 N_0$. Instead, the dependence is on the operator norm of the empirical second moment matrix $\norm{\bSigma_N}{2}$, and polynomial in $X_*$, $W_*$, $\log(1/\delta)$, and $1/\eps$. This is favorable, as in many settings the total number of pairs can be prohibitively large. The following is the main result of this section, regarding the actual risk of LCBR solution.

\textbf{Theorem 3.~} For a probability target $p^{\star}$ let the sample size be chosen such that
\begin{align}\label{eq:subsample_comp2}
S \geq \max \biggl\{ \log(4D/p^{\star}) \frac{\norm{\bSigma_N}{2}X_*^2 W_*^4 + (1/15)\eps X_*^2 W_*^2}{\eps^2/800} ~,~ \frac{X_*^2 W_*^2}{\eps^2/100} \brac{\sqrt{2\log(4/p^\star)}+1}^2 \biggr\} ~.
\end{align}
Then the solution $\bw_S$ returned by LCBR satisfies
\begin{align}\label{eq:lcbr_regret}
\underset{\cN \sim \cD^N}{P} \bigg( R(\bw_S)-R(\bw_{\star}) \geq \eps \bigg) \leq 2 ~ C_{\cS}\paren{\frac{\eps}{10C_1}} ~ \exp \Brac{-\frac{\eps^2}{50 C_2^2}\rho (1-\rho) N} + p^{\star} ~.
\end{align}

\textbf{Proof.~} We start with the inequality 
\begin{align}
R(\bw_S)-R(\bw_{\star}) = \magnitude{R(\bw_S)-R(\bw_{\star})} \leq \magnitude{R(\bw_S)-R(\bw_N)} + \magnitude{R(\bw_N)-R(\bw_{\star})}.
\end{align}
The first term can be bounded as
\begin{align} \label{eq:thm3_ineq1}
\magnitude{R(\bw_S)-R(\bw_N)} &= \magnitude{R(\bw_S)-\RN(\bw_S)+\RN(\bw_S)-\RN(\bw_N)+\RN(\bw_N)-R(\bw_N)} \nn \\
&\leq \magnitude{R(\bw_S)-\RN(\bw_S)} + \magnitude{R(\bw_N)-\RN(\bw_N)} + \magnitude{\RN(\bw_S)-\RN(\bw_N)} ~.
\end{align}
Note that the empirical risk trick in the proof Theorem 1 no longer applies as it not known which one of $R(\bw_N)$ or $R(\bw_S)$ is smaller, and all three terms have to be retained. On the other hand, for the second term, from the proof of Theorem 1 we know that
\begin{align} \label{eq:thm3_ineq2}
\magnitude{R(\bw_N)-R(\bw_\star)} \leq \magnitude{R(\bw_N)-\RN(\bw_N)} + \magnitude{R(\bw_\star)-\RN(\bw_\star)} ~.
\end{align}
Combining Eqs.\ \eqref{eq:thm3_ineq1} and \eqref{eq:thm3_ineq2} we get
\begin{align}
R(\bw_S) - R(\bw_\star) \leq \bigg[ \sum_{\bw \in \cW_4} \magnitude{R(\bw)-\RN(\bw)} \bigg] + \magnitude{\RN(\bw_S)-\RN(\bw_N)} ~.
\end{align}
where we defined the sequence $\cW_4 = [\bw_{\star},\bw_N,\bw_N,\bw_S]$. We now consider bounding each term by $\eps/5$. For the summation on the right hand side this yields
\begin{align}
P \paren{\sum_{\bw \in \cW_4} \magnitude{R(\bw)-\RN(\bw)} \leq 4\eps/5} &\leq P \paren{\sup_{\bw \in \cW} \magnitude{R(\bw)-\RN(\bw)} \leq \eps/5} \nn \\
&\leq 2 ~ C_{\cS}\paren{\frac{\eps}{10 C_1}} ~ \exp \Brac{-\frac{\eps^2}{50 C_2^2}\rho (1-\rho) N}
\end{align}
where the last inequality follows from Lemma 1. For the last term we would like to bound the deviation by $\eps/5$ with probability $p^{\star}$. Plugging these terms into the sample complexity bound of Theorem 2 yields the bound of Eq.\ \eqref{eq:subsample_comp2}. $\hfill \square$

Theorem 3 shows that the number of samples $S$ required to make LCBR competitive with BBR does not depend on $N_1N_0$. Firstly, for a fixed probability target $p^\star$, $S$ depends on $\cN$ only through $\bSigma_N$. In practice we set $p^\star$ such that it is relatively smaller than the exponential term in Eq.\ \eqref{eq:lcbr_regret}. In this case $S$ will have an implicit dependence on $N_1+N_0$ and $\rho$. The important difference is, while BBR requires $N_1N_0$ samples to achieve the bound in Eq.\ \eqref{eq:bbr_regret}, LCBR can achieve the comparable bound in Eq.\ \eqref{eq:lcbr_regret} with $S \ll N_1N_0$. We demonstrate this with experiments in Section \ref{sec:exp}. Finally, note that we focused on the linear case due to space limitations. Clearly, for any \emph{fixed} nonlinear feature map, our results hold where $D$ is replaced with the dimensionality of the new feature space. Another direction is to consider random feature transforms and provide bounds based to the best ranker in the corresponding function space \cite{Recht_2008}; we leave this for a longer version of the paper.

\section{Related Work}
\label{sec:rel}

Our proof of the uniform risk bound in Lemma 1 is based on covering numbers, which is also used for analyzing linear regression problems \cite{Mohri_2012}. The covering number-based argument is later extended to online learning with pairwise loss functions \cite{Wang_2012}; however, their analysis is for sequential updates. Bounds based on Rademacher complexity and U-processes are considered in \cite{Kar_2013, Clemencon_2008, Peel_2010, Lei_2018}. These bounds can be tighter, however they are based on the assumption that all samples are drawn i.i.d. from $\cD$, which is different from our setup. Bounds based on algorithmic stability \cite{Agarwal_2009, Cortes_2007} and VC dimenstion \cite{Agarwal_2005} have also been considered. Replacing the discrete AUC loss with a convex one has been investigated in a number of work. In particular \cite{Zhao_2011, Gao_2013, Kar_2013, Ding_2015, Boissier_2016} consider online algorithms. Online learning based on stochastic saddle point problems is considered in \cite{Natole_2018, Ying_2016}. It is also worthwhile to note that the majority of the aforementioned papers are based on the pairwise squared loss. In \cite{Gao_2015}, the consistency of surrogate loss functions with respect to AUC loss has been considered, extending the results of \cite{Bartlett_2006}. On the other hand \cite{Agarwal_2014, Kotlowski_2011} provide bounds for the AUC loss in terms of the surrogate loss of the learner. Kernel based methods for bipartite ranking was considered in \cite{Hong_2007, Ding_2017, Hu_2015}. (For more related work see also \cite{Narasimhan_2013, Agarwal_2008, Rajkumar_2016}.) In terms of low sample complexity, the stochastic gradient descent (SGD) based online learning algorithms also provide bounds in terms of the samples used; however these algorithms require a step size to tune, whereas LCBR in Algorithm 2 does not need this parameter, which can be an important practical advantage. As we will show in experiments, LCBR can achieve better performance with fewer samples, compared to the SGD-based methods.

\section{Experiments}
\label{sec:exp}

\subsection{Gaussian Mixture Distribution}

\begin{figure}[t]
	\centering
	\begin{subfigure}{.32\columnwidth}
		\centering
		\includegraphics[scale=.3]{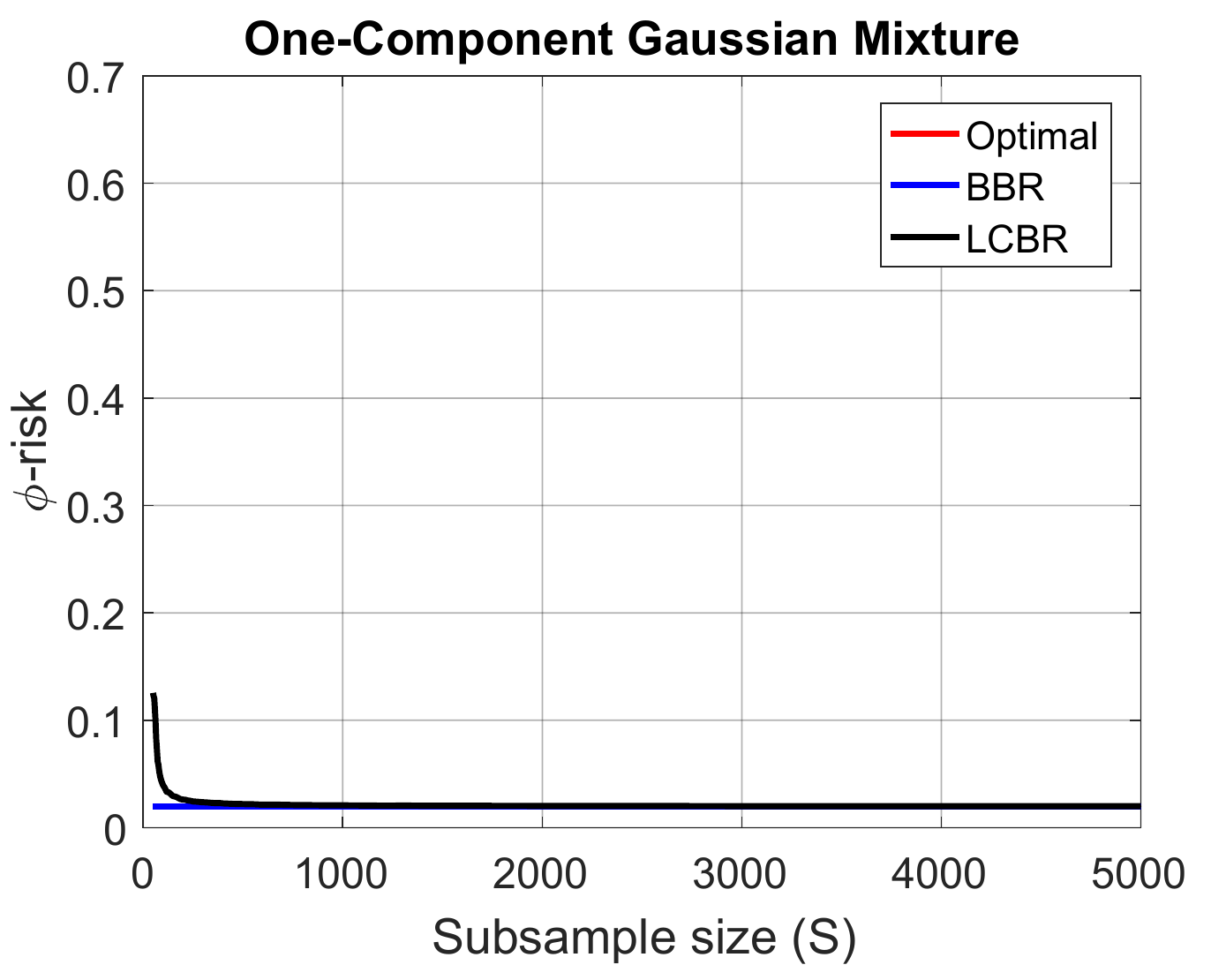}
		\caption{}
	\end{subfigure} %
	\begin{subfigure}{.32\columnwidth}
		\centering
		\includegraphics[scale=.3]{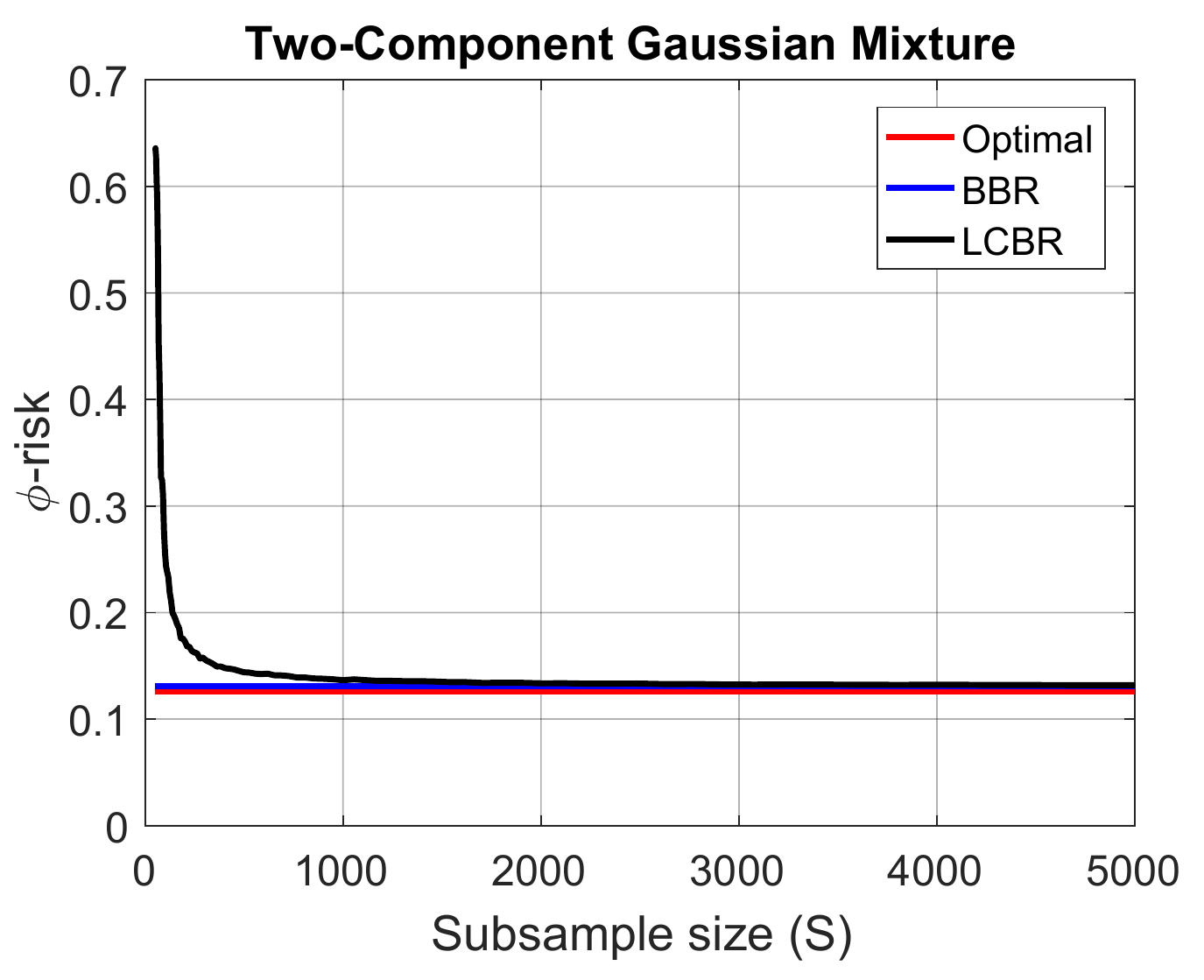}
		\caption{}
	\end{subfigure} %
	\begin{subfigure}{.32\columnwidth}
		\centering
		\includegraphics[scale=.3]{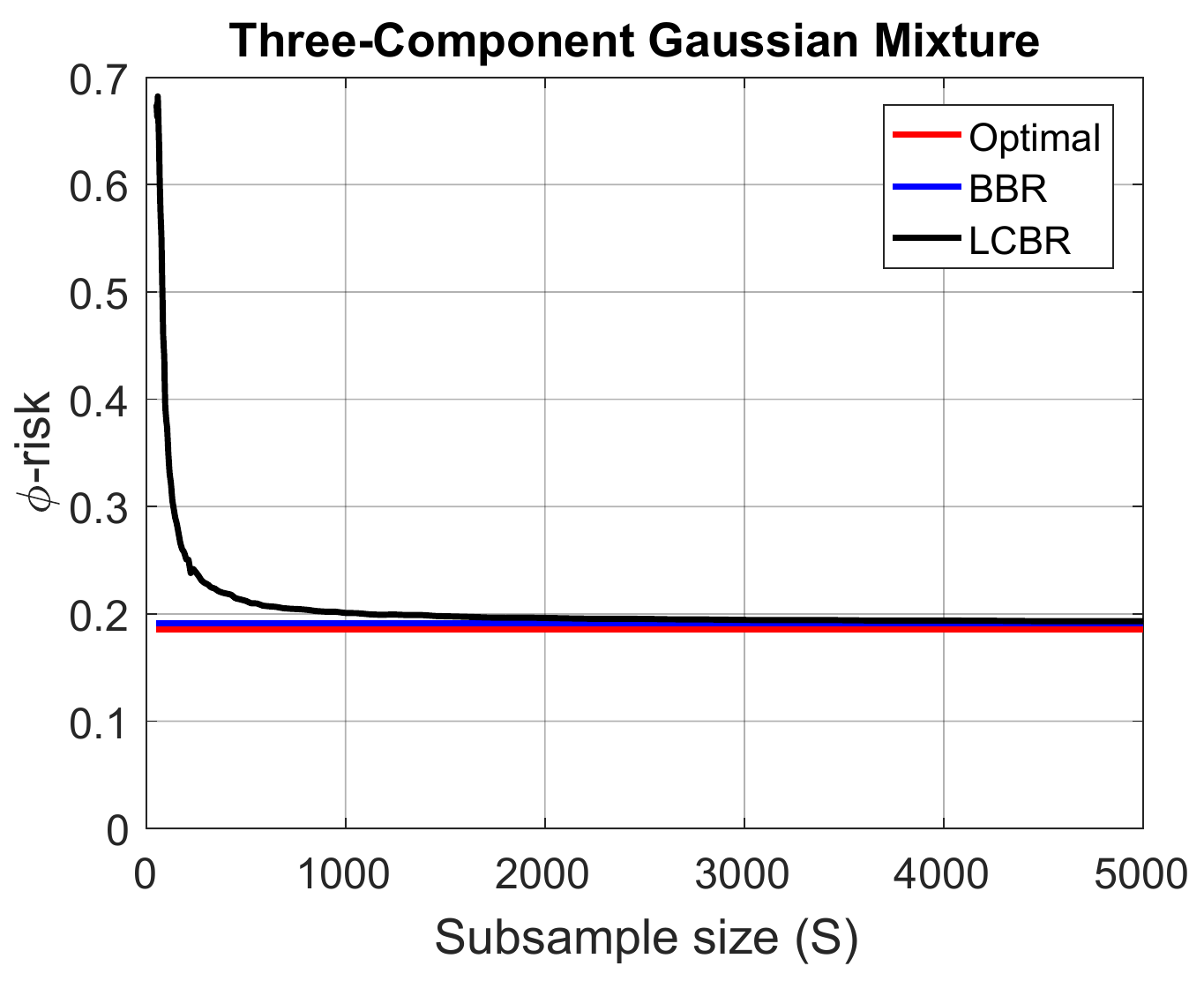}
		\caption{}
	\end{subfigure} %
	\begin{subfigure}{.32\columnwidth}
		\centering
		\includegraphics[scale=.3]{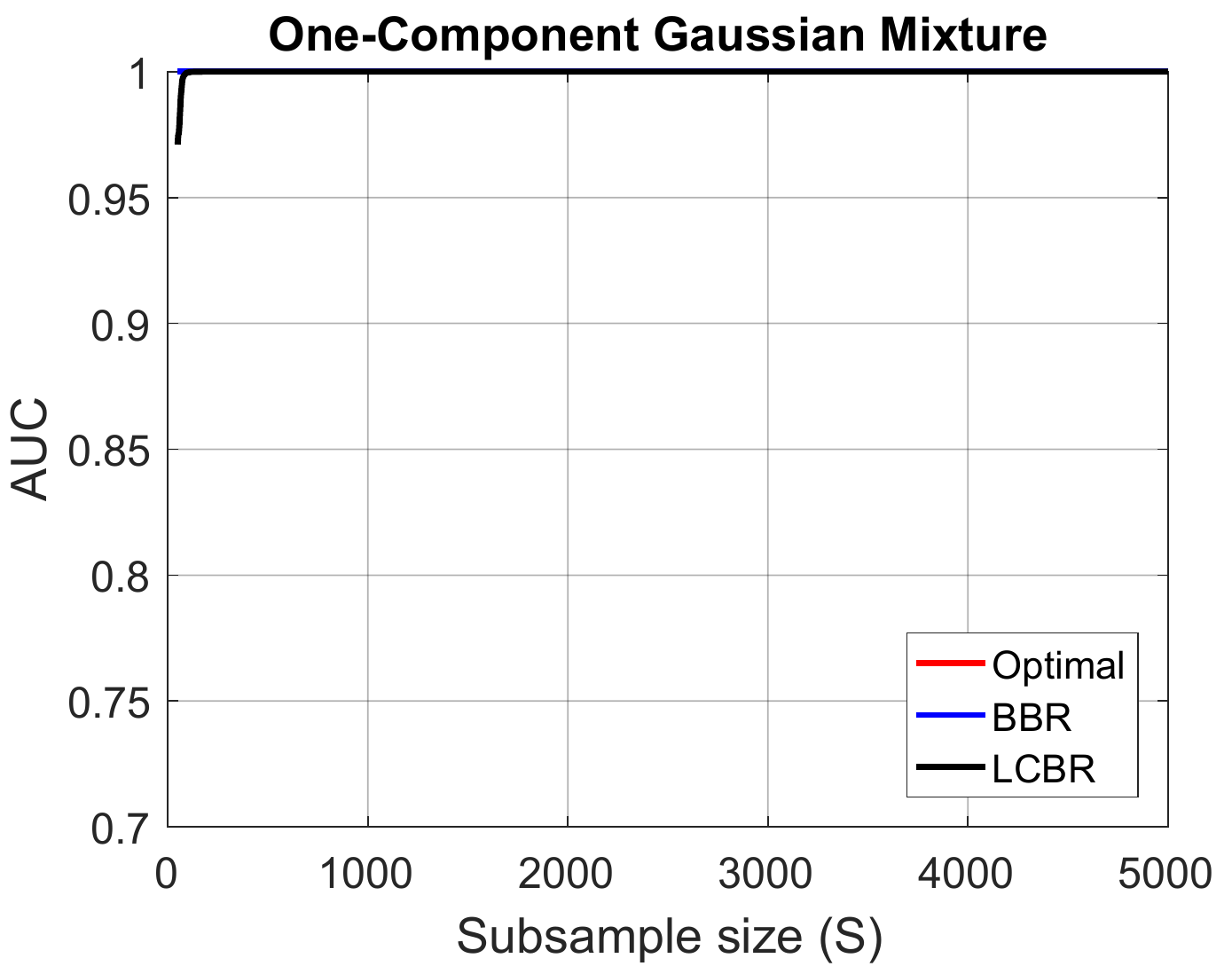}
		\caption{}
	\end{subfigure} %
	\begin{subfigure}{.32\columnwidth}
		\centering
		\includegraphics[scale=.3]{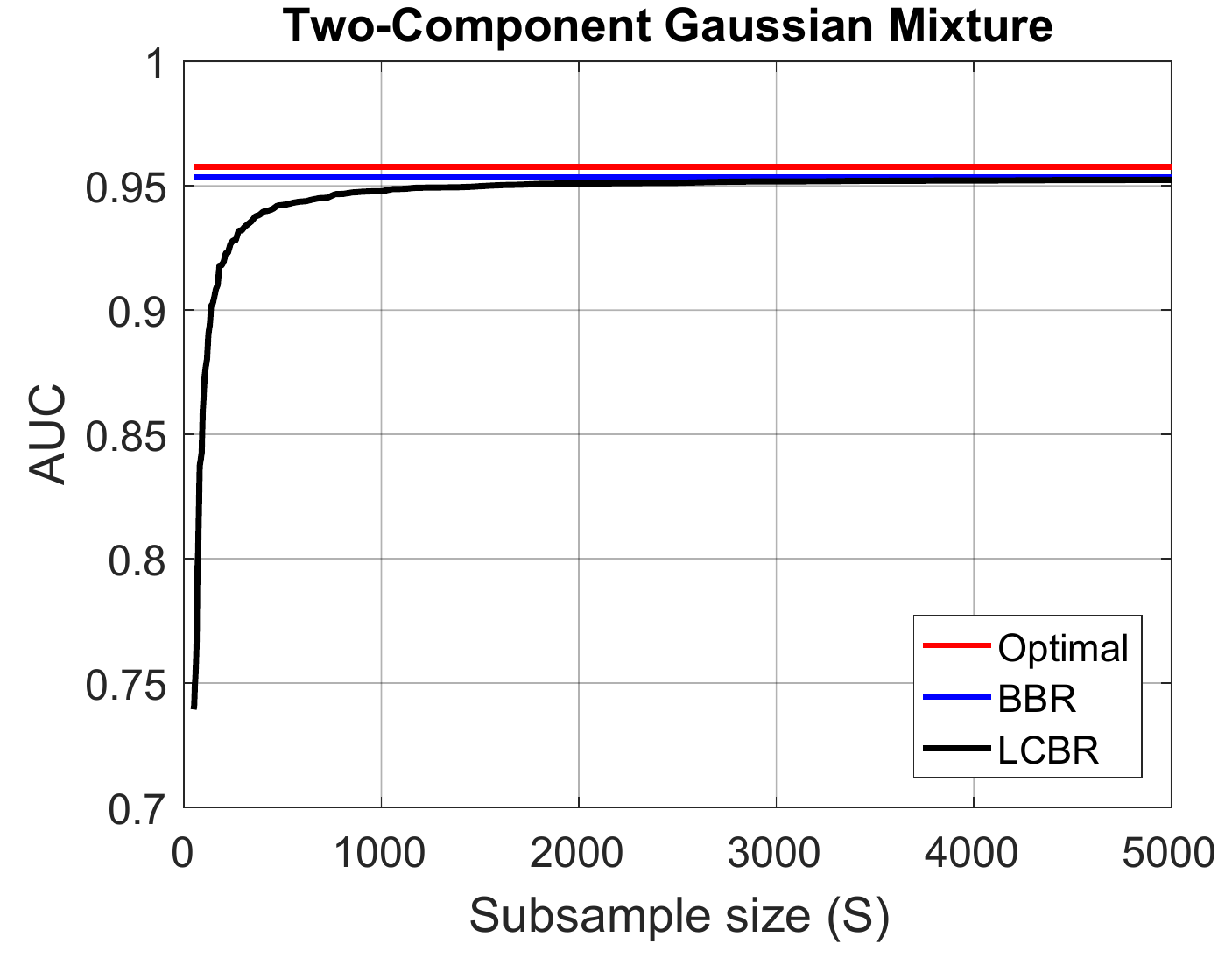}
		\caption{}
	\end{subfigure} %
	\begin{subfigure}{.32\columnwidth}
		\centering
		\includegraphics[scale=.3]{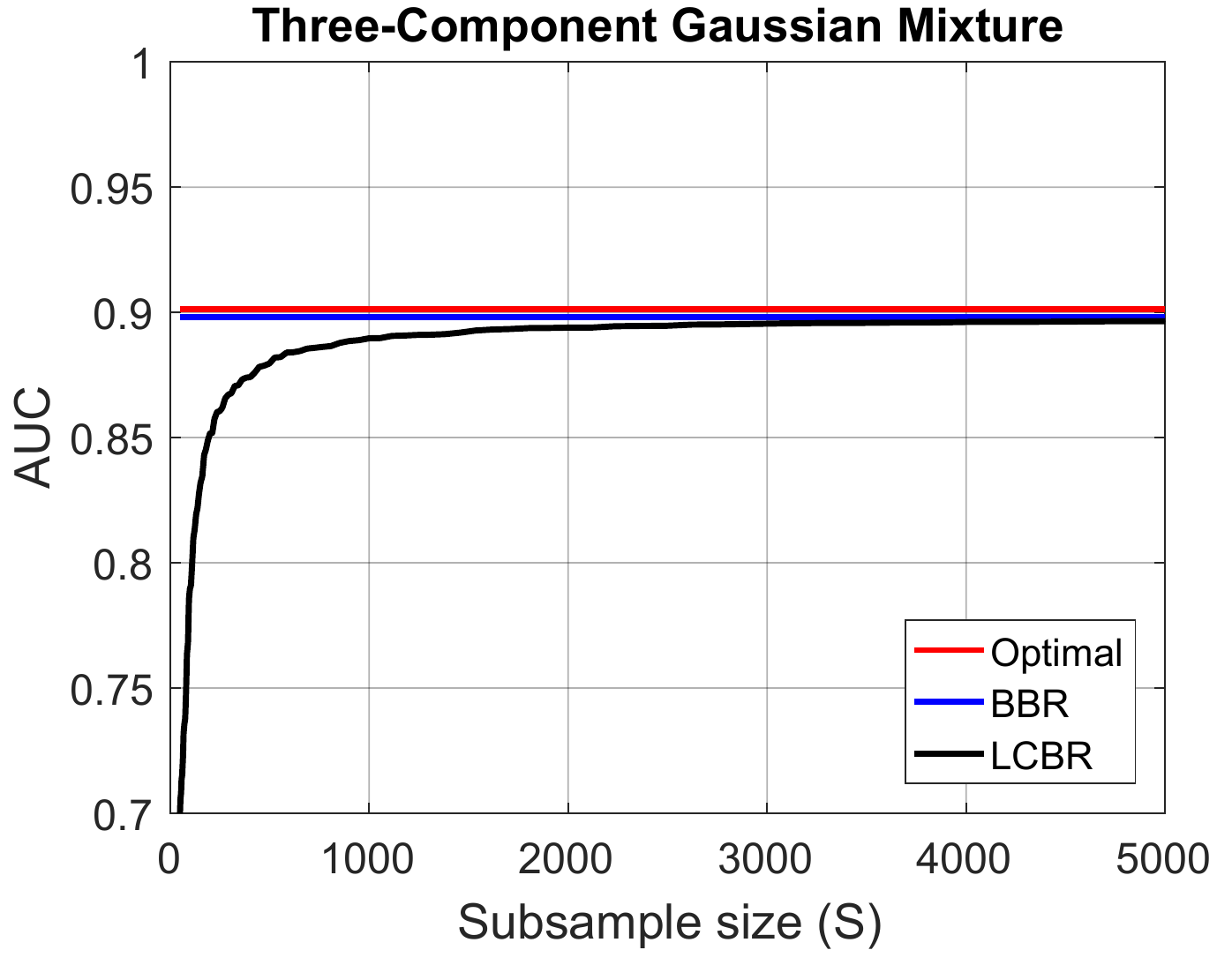}
		\caption{}
	\end{subfigure} %
	\begin{subfigure}{.2\columnwidth}
		\centering
		\includegraphics[scale=.35]{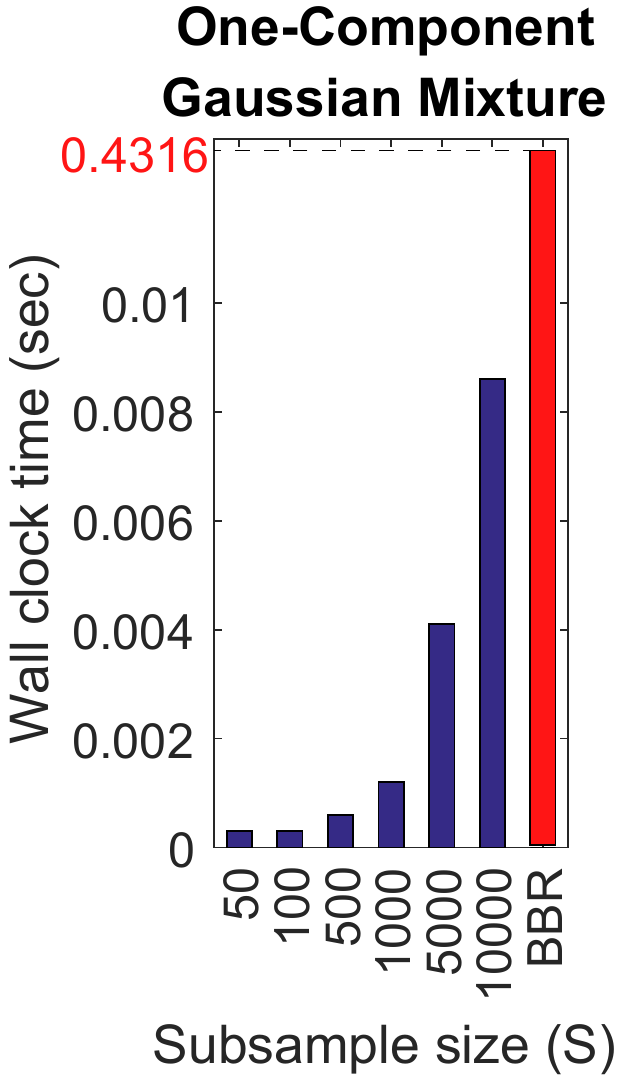}
		\caption{}
	\end{subfigure} %
	\begin{subfigure}{.2\columnwidth}
		\centering
		\includegraphics[scale=.35]{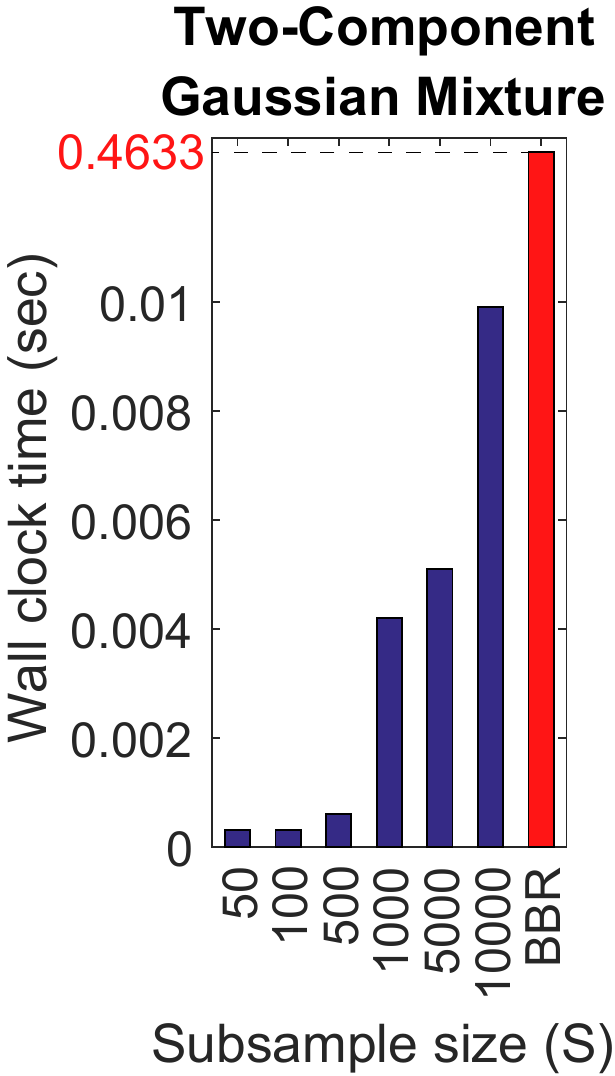}
		\caption{}
	\end{subfigure} %
	\begin{subfigure}{.2\columnwidth}
		\centering
		\includegraphics[scale=.35]{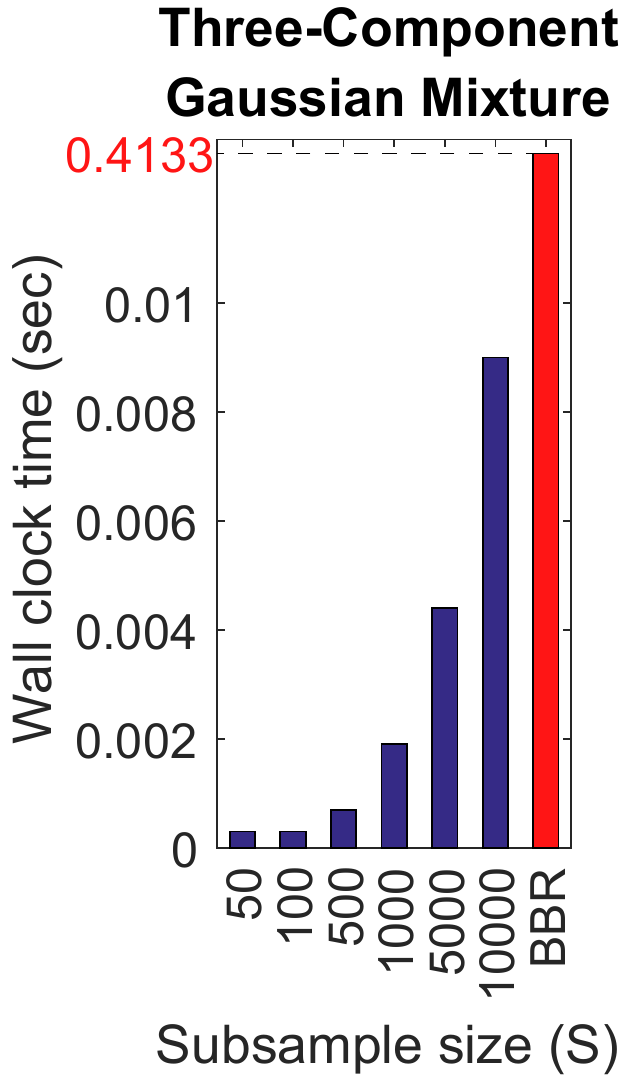}
		\caption{}
	\end{subfigure} %
	\begin{subfigure}{.32\columnwidth}
		\centering
		\includegraphics[scale=.35]{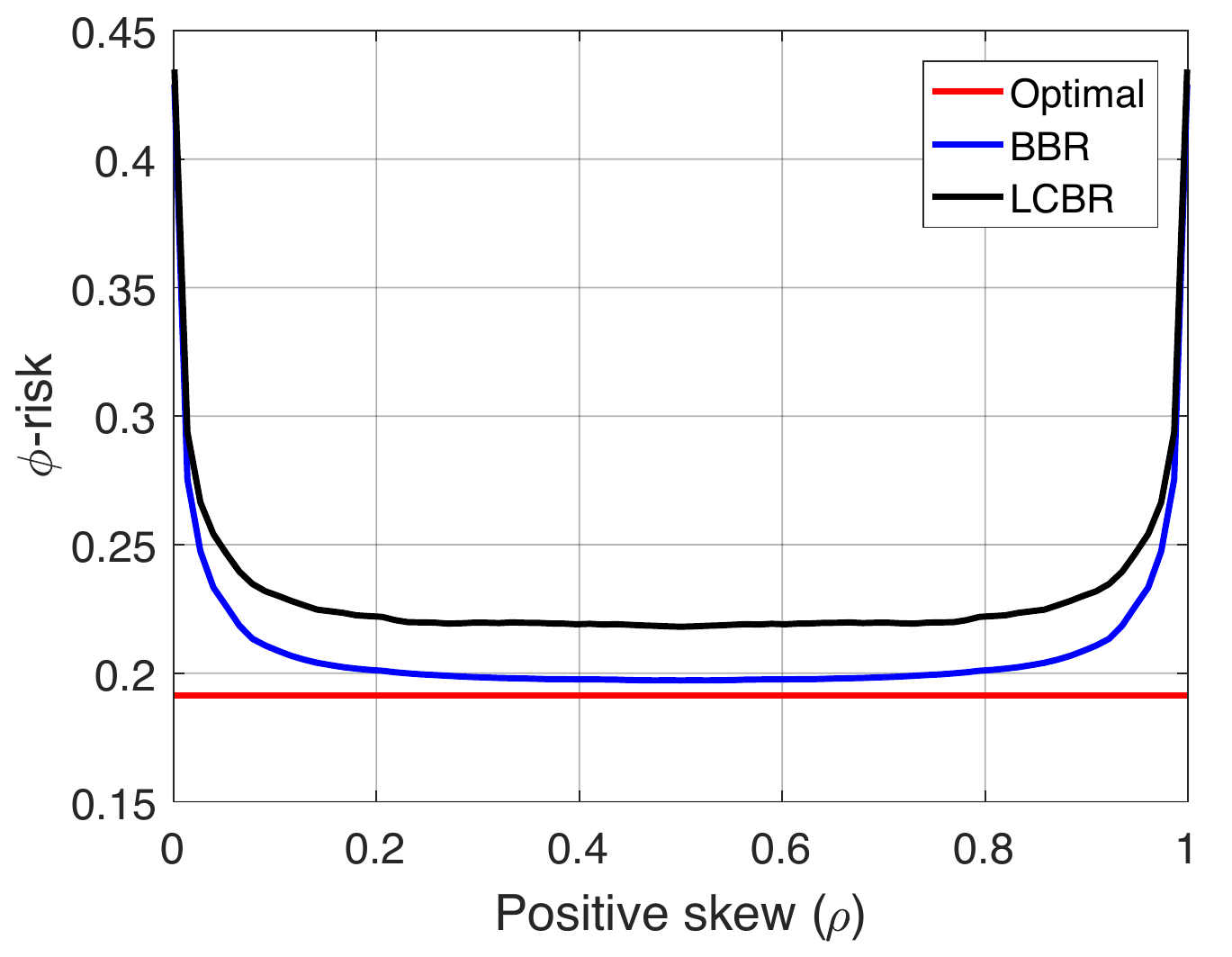}
		\caption{}
	\end{subfigure} %
	\caption{Results of experiments with Gaussian Mixture distributions.}
\end{figure}

We first consider experiments where the generating distribution is a $K$-component Gaussian Mixture. So for $i\in\set{0,1}$ we can write
\begin{align}
\cD^i(\bx) = \sum_{k=1}^K \frac{c_k}{\sqrt{2\pi\sigma^2}} \exp \{-\norm{\bx - \bmu_k^i}{2}^2 / (2\sigma^2) \}
\end{align}
where $c_k$'s are mixture weights. We consider the case where the covariance matrix is isotropic and controlled with a single scale parameter $\sigma$. We consider three cases where $K=1,2,3$ and $\sigma=2,3,4$ where the increasing value of $K$ and $\sigma$ makes the problem gradually more difficult. The mean values for class one are sampled from the unit cube in positive orthant and class zero from unit cube in negative orthant. The curves are obtained by averaging 50 experiments. For any given pair of conditional distributions the optimal ranking function minimizing the $\phi$-risk is found by calculating $\bmu$ and $\bSigma$ and solving the $\phi$-risk minimization problem. On the other hand, the optimal ranking function maximizing the AUC is the likelihood ratio of class-conditionals, which is a consequence of the Neyman-Pearson lemma \cite{Poor_1998}. Figure 1 shows the results of our experiments. For panels (a)-(f) we have $\rho=0.5$ with $10^3$ positive and negative samples. So BBR uses $10^6$ pairs whereas LCBR uses the number shown on the x-axis. In panels (a)-(c) we show the $\phi$-risk and in (d)-(f) we show the AUC. As the number of components increase the problem becomes more difficult, resulting in higher $\phi$-risk and lower AUC for optimal ranking functions. In all cases BBR is very close to the optimal ranker; but it has high sample cost. LCBR, on the other hand, catches on with $S = 3000$ at most. This is also reflected in wall clock times; as shown in panels (g)-(i) LCBR is roughly 100 times faster than BBR when $S=5000$ subsamples are used. Finally we illustrate the performance of BBR and LCBR as a function of label skew $\rho$. In the proof of Lemma 1, we saw that applying McDiarmid's inequality gives the term $\rho(1-\rho)$ in the exponent. This suggests, as $\rho$ goes to $0$ or $1$ the generalization performance should decrease. Panel (j) shows that this is indeed the case: here BBR uses $2000\rho$ positive and $2000(1-\rho)$ negative samples, whereas LCBR uses $5,000$ subsamples. The performance of BBR and LCBR are once again close; but as $\rho$ deviates from $0.5$ the generalization of both algorithms get worse.

\subsection{LIBSVM Datasets}

We now compare LCBR against state-of-the art algorithms on three datasets from LIBSVM. The algorithms we implement, in addition to LCBR, are the following: AdaOAM \cite{Ding_2015} uses adaptive stochastic gradient descent with pairwise squared loss, whereas PGD is SGD based approach \cite{Boissier_2016}. SPAM is one of the most recent works where AUC optimization is formulated as a stochastic saddle point problem \cite{Natole_2018}. OAM is a relatively older algorithm, but we include it as it is based in a different objective, the pairwise hinge loss \cite{Zhao_2011}. As widely done in the literature, we use AUC on test set as the performance measure. We use the train and test splits provided by LIBSVM. Regularization parameters are determined by cross-validation and step sizes are chosen based on the references. The experiments are averaged over 50 runs. 

We show the results in Figure 2. The x-axis correspond to the number of subsamples used as a percentage of total number of samples available. For example, {\tt A9A} dataset contains approximately $33K$ data points. A subsample ratio of $50\%$ implies we use approximately $16.5K$ random samples from this dataset. For the {\ttfamily A9A} dataset all algorithms have good generalization, with an AUC of approximately $90\%$. For the {\tt GERMAN} and {\tt SVMGUIDE3} datasets, on the other hand, there is a significant gap between LCBR and the SGD-based competitiors. Here the SGD and learning rate free approach of LCBR proves useful and it achieves better generalization for the given number of samples. Therefore the sample complexity of LCBR is favorable.

\subsection{Algorithmic Complexity}

For the algorithms presented in this paper, it can be seen that the computational bottleneck is at the accumulation phase (cf. Algorithms 1 and 2). For BBR the cost of this step is $O(N_1 N_0 D^2)$ and for LCBR this is $O(S D^2)$. Typically $S \ll N_1 N_0$ which can save significant computation. On the other hand, the storage for both BBR and LCBR is $O(D^2)$ as it requires storing the covariance matrix. This quadratic dependence on dimension can be an issue when $D$ is too large. In this case, a sparse approximation of $\bSigma_S$ can be necessary. However, note that this storage bottleneck is also present in the algorithms we compared \cite{Zhao_2011,Boissier_2016,Ding_2015}. In contrast, SPAM \cite{Natole_2018} can be implemented in $O(D)$ space; however that algorithm requires knowledge of expectations with respect to the true conditionals, which is unknown in practice. Our experiments show that LCBR can still achieve better performance.

\begin{figure}[t]
	\centering
	\begin{subfigure}{.32\columnwidth}
		\centering
		\includegraphics[scale=.33]{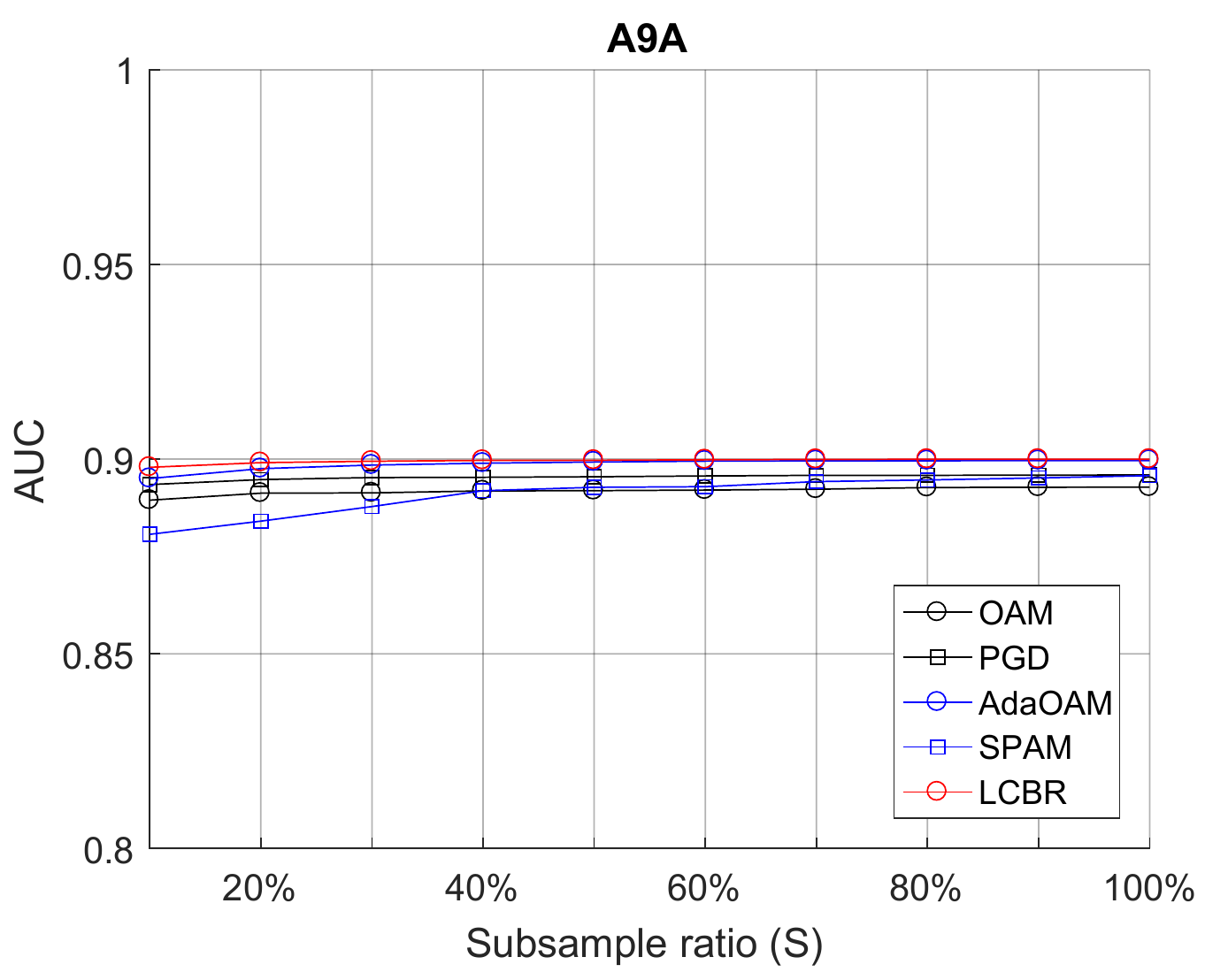}
		\caption{}
	\end{subfigure} %
	\centering
	\begin{subfigure}{.32\columnwidth}
		\centering
		\includegraphics[scale=.33]{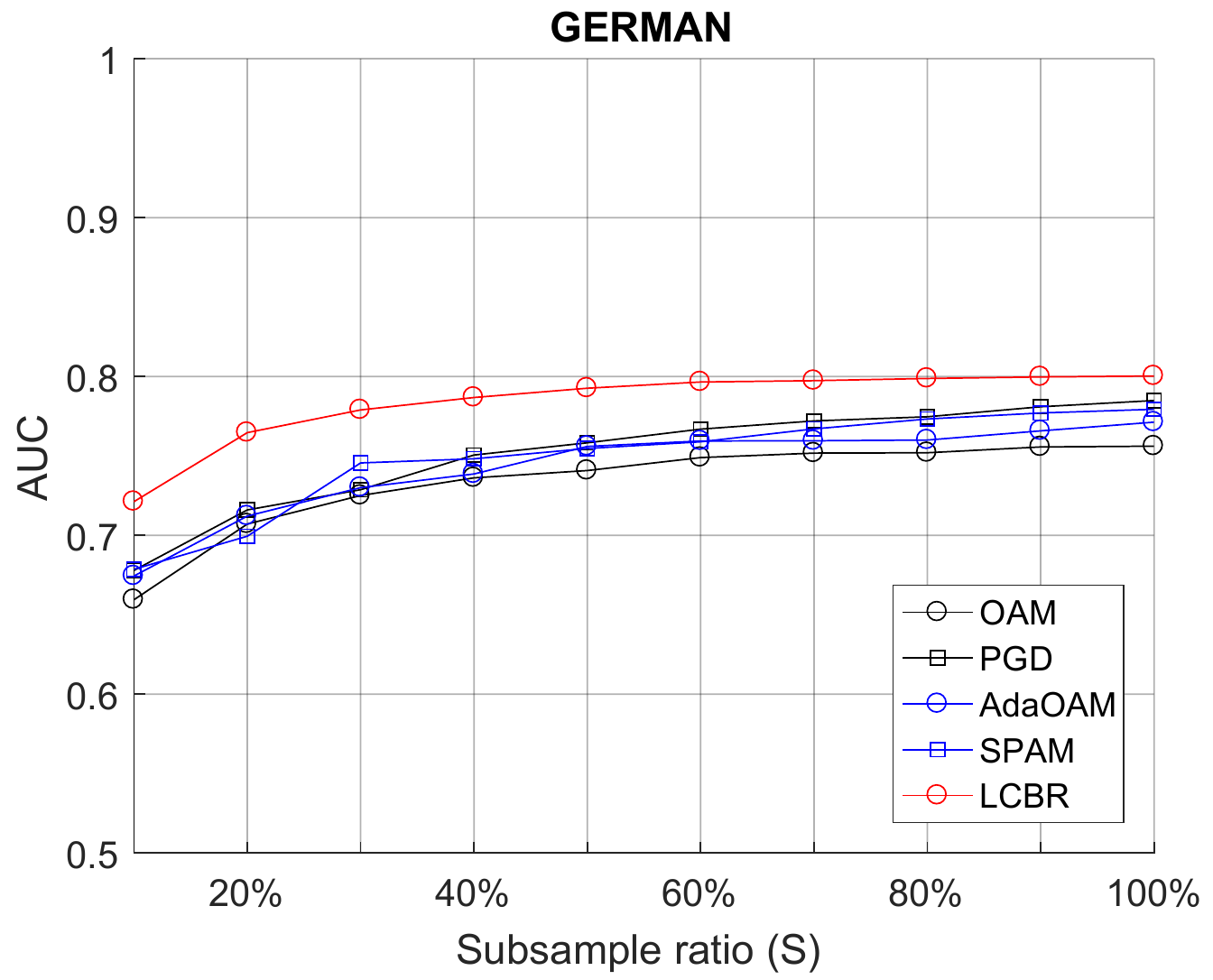}
		\caption{}
	\end{subfigure} %
	\centering
	\begin{subfigure}{.32\columnwidth}
		\centering
		\includegraphics[scale=.33]{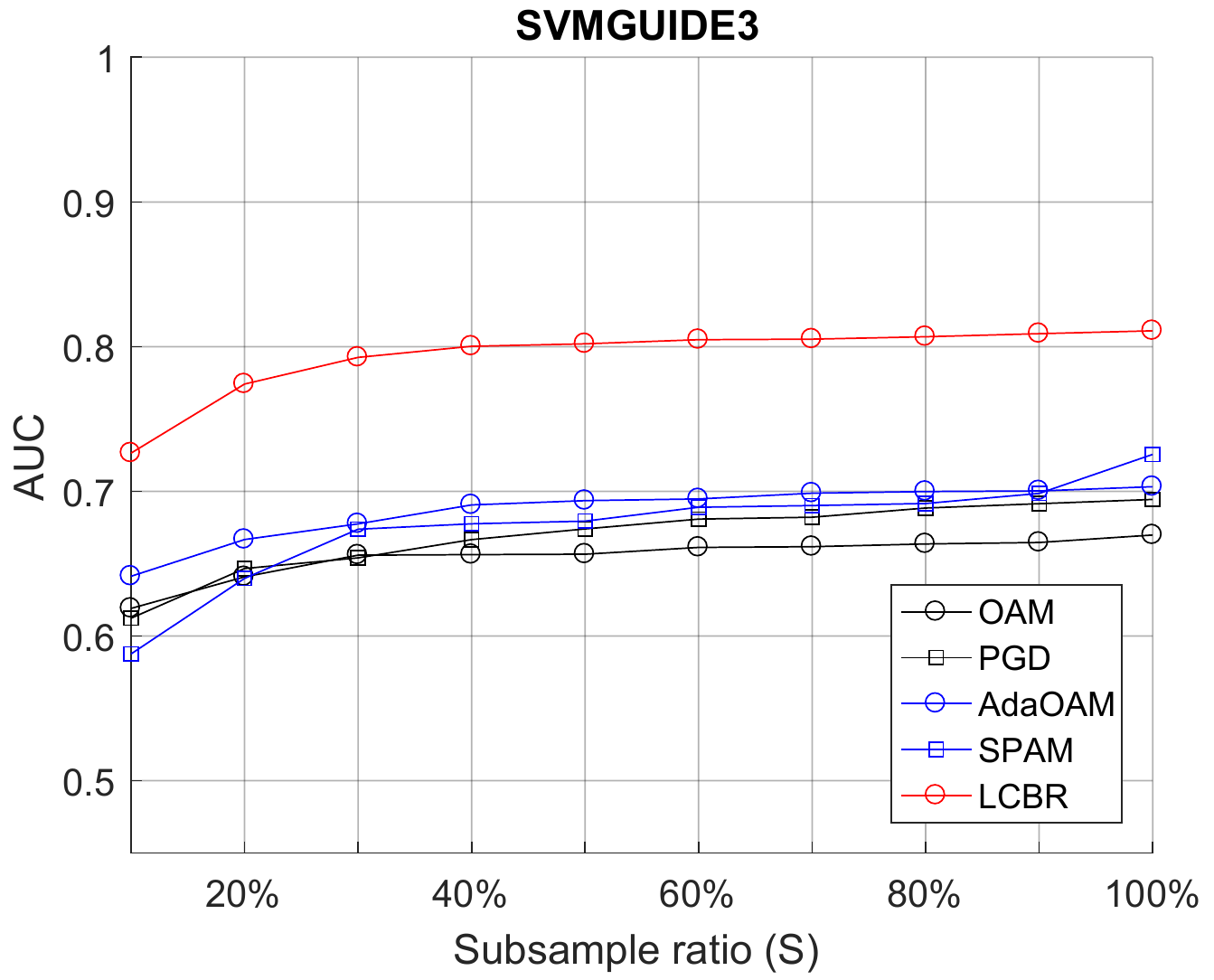}
		\caption{}
	\end{subfigure} %
	\caption{Results of experiments with LIBSVM datasets.}
\end{figure}

\section{Conclusion}
\label{sec:con}

We have considered the problem of bipartite ranking, where the empirical AUC loss is replaced with the pairwise squared loss. Different from the previous work---which was based on SGD---we proposed a low sample cost bipartite ranking algorithm (LCBR), and showed that the number of samples required for good performance obeys $S \ll N_1 N_0$. Experiments show that LCBR quickly achieves similar performance with BBR where the number of samples are several order of magnitudes lower. Experiments against state-of-the-art bipartite ranking algorithms also show that LCBR can achieve better generalization with a smaller subsample set. In a longer version of the paper we will also consider extending these results to random feature spaces, which include random kernel features and random neural networks.

\section{Appendix: Proof of Lemma 2}

For the proofs we will use the shorthand $\bx_s=\bx_{i_s}^1 - \bx_{j_s}^0$.

(i) We recall the Matrix Bernsten Inequality for a $D \times D$ symmetric, random matrix $\bZ$ and threshold $\gamma$:
\begin{align} \label{eq:conc_matrix}
P(\norm{\bZ}{2} > \gamma) \leq 2D \exp \paren{-\frac{\gamma^2/2}{\bV(\bZ) + L\gamma/3}}
\end{align}
where $L$ is a norm bound on summands. 

First apply a spectral norm bound to the quadratic expression: $\sup_{\bw_1 \in \cW} \magnitude{\DS} \leq W_*^2 \norm{\bSigma_S-\bSigma_N}{2}$. We now take $\bZ = \bSigma_S-\bSigma_N$. The spectral norm can be bounded based on the argument in \cite{Tropp_2015}. We can decompose $\bZ$ into a sum: $\bZ = \sum_{s=1}^S (1/S) [\bx_s \bx_s^\top - \bSigma_N]$. We denote each summand by $\bE_s = (1/S)[\bx_s \bx_s^\top - \bSigma_N]$. It then follows from triangle inequality that
\begin{align} 
\norm{\bE_s}{2} \leq \frac{1}{S} \brac{\norm{\bx_s\bx_s^\top}{2} + \norm{\bSigma_N}{2}} \leq \frac{2}{S} \norm{\bx_s}{2}^2 \leq \frac{8X_*^2}{S}
\end{align}
where the second inequality follows from Jensen's inequality. Since the subsampling is conditional on $\cN$, $\bSigma_N$ is constant with respect to the subsampled pairs, and each summand is centered and i.i.d. The variance of the sum decomposes as $\bV(\bZ) = \norm{\sum_{s \in S}\EX[\bE_s^2]}{2}$. For a single summand the second moment can be bounded as
\begin{align}
\EX[\bE_s^2] = \frac{1}{S^2} \EX\brac{\brac{\bx_s\bx_s^\top - \bSigma_N}^2} = \frac{1}{S^2} \brac{\EX\brac{\norm{\bx_s}{2}^2 \bx_s \bx_s^\top} - \bSigma_N^2} \preceq \frac{4 X_*^2}{S^2} \bSigma_N
\end{align}
from which the variance inequality $\bV(\bZ) \leq (4 X_*^2 / S) \norm{\bSigma_N}{2}$ follows. Substituting these to Eq.\ \eqref{eq:conc_matrix} with $\gamma = \eps/W_*^2$ yields the result.

(ii) We recall the following concentration inequality for i.i.d. and bounded random vectors with mean $\bar{\bx}$ \cite{Recht_2008}: 
\begin{align} \label{eq:conc_vector}
P \paren{ \norm{\frac{1}{S}\sum_{s=1}^S \bx_s - \bar{\bx}}{2} \geq \gamma } \leq \exp \Brac{-\frac{1}{2}\paren{\frac{\gamma \sqrt{S}}{L}-1}^2} ~.
\end{align}
From the following inequalities
\begin{align}
\sup_{\bw_1,\bw_2 \in \cW} \magnitude{\Ds(\bw_1,\bw_2)} \leq \sup_{\bw_1,\bw_2} \norm{\bw_1-\bw_2}{S} \norm{\bmu_N-\bmu_S}{2} \leq 2 W_* \norm{\frac{1}{S}\sum_{s=1}^S \bx_s - \bmu_N}{2}
\end{align}
the desired result is obtained by setting $\gamma = \eps/2 W_*$ and $L=X_*$ in Eq.\ \eqref{eq:conc_vector}. $\hfill \square$

\bibliographystyle{IEEEtran}
\bibliography{lcbr_bib}	
	
\end{document}